\documentclass[runningheads]{llncs}
\usepackage[T1]{fontenc}
\usepackage{graphicx}
\usepackage{graphicx}
\usepackage{cite}
\usepackage{subcaption}
\usepackage{amsmath}
\usepackage{amssymb}
\usepackage{booktabs}
\usepackage{float}
\usepackage[pagebackref,breaklinks,colorlinks]{hyperref}
\usepackage[capitalize]{cleveref}
\usepackage{colortbl}
\usepackage[table]{xcolor}

\crefname{section}{Sec.}{Secs.}
\Crefname{section}{Section}{Sections}
\Crefname{table}{Table}{Tables}
\crefname{table}{Tab.}{Tabs.}

\begin{document}
\title{Weakly Supervised Intracranial Hemorrhage Segmentation with YOLO and an Uncertainty Rectified Segment Anything Model}
\titlerunning{Uncertainty-Rectified YOLO-SAM for Weakly Supervised ICH Segmentation}
%
\author{Pascal Spiegler\inst{1}\index{Spiegler, Pascal} \and
Amirhossein Rasoulian\inst{2}\index{Rasoulian, Amirhossein} \and
Yiming Xiao\inst{1}\index{Xiao, Yiming}}
\authorrunning{P. Spiegler et al.}
%
\institute{Department of Computer Science and Software Engineering, Concordia University, Montreal, Canada 
\email{\\pascal.spiegler@mail.concordia.ca}\\
\and
NeuroRx Research, Montreal, Canada}
\maketitle              
\begin{abstract}
Intracranial hemorrhage (ICH) is a life-threatening condition that requires rapid and accurate diagnosis to improve treatment outcomes and patient survival rates. Recent advancements in supervised deep learning have greatly improved the analysis of medical images, but often rely on extensive datasets with high-quality annotations, which are costly, time-consuming, and require medical expertise to prepare. To mitigate the need for large amounts of expert-prepared segmentation data, we have developed a novel weakly supervised ICH segmentation method that utilizes the YOLO object detection model and an uncertainty-rectified Segment Anything Model (SAM). In addition, we have proposed a novel point prompt generator for this model to further improve segmentation results with YOLO-predicted bounding box prompts. Our approach achieved a high accuracy of 0.933 and an AUC of 0.796 in ICH detection, along with a mean Dice score of 0.629 for ICH segmentation, outperforming existing weakly supervised and popular supervised (UNet and Swin-UNETR) approaches. Overall, the proposed method provides a robust and accurate alternative to the more commonly used supervised techniques for ICH quantification without requiring refined segmentation ground truths during model training.

\end{abstract}
\section{Introduction}

Intracranial hemorrhage (ICH) accounts for 10-15\% of all stroke cases and carries a significant risk of mortality \cite{StrokeStats}. Hemorrhage volume, which can rapidly expand within the first few hours, is a key predictor of treatment outcomes and potential complications \cite{VolumeCorrelation}. Precise localization and quantification of the five ICH subtypes, including intraventricular (IVH), intraparenchymal (IPH), subarachnoid (SAH), epidural (EDH), and subdural (SDH), are therefore essential for tailoring treatment strategies and minimizing adverse events \cite{Subtypes}. While supervised deep learning (DL) models have demonstrated excellent potential in automating ICH assessment \cite{heit2021automated}, their success heavily relies on large datasets with pixel-level annotations (ground-truth masks) and poor segmentation accuracy is observed with smaller training datasets \cite{CNNSegmentation}. However, large training datasets containing high-quality ground-truth masks are difficult to obtain due to high demands in time, labor, and domain expertise. Together with scarce public ICH segmentation datasets, this bottleneck poses great challenges in developing automatic ICH quantification algorithms to better facilitate the care and management of the condition.

To overcome the aforementioned issue, weakly supervised learning approaches \cite{rasoulian-2023, 2022Rasoulian} have emerged as a promising alternative. These methods leverage more economic ground truths, such as categorical labels, bounding boxes, or coarse masks to train segmentation models, bypassing the requirement of refined masks for fully supervised and semi-supervised approaches. While most existing literature is dedicated to ICH detection, ICH segmentation using weakly supervised methods remains under-explored. However, limited prior explorations exist leveraging explainable AI methods for weakly-supervised stroke segmentation, including class-activation maps (CAM) \cite{wu2019weakly} and self-attention maps \cite{rasoulian-2023}, providing encouraging results. Recent developments in foundation models, such as the Segment Anything Model (SAM) \cite{SAM} have shown great potential to mitigate the segmentation ground truth bottleneck, but have not been explored for improving weakly supervised ICH segmentation. Therefore, we propose a novel weakly supervised ICH segmentation technique that incorporates automatic box and point prompt generation with SAM to allow for ICH detection and segmentation on CT scans. We have three main contributions. \textbf{First}, we leveraged a finetuned YOLOv8 model and a novel morphology-based method to automatically generate box and point prompts, respectively, for SAM. \textbf{Second}, to enhance segmentation accuracy with SAM, we employed an uncertainty rectification approach to account for uncertainty in prompt generation. \textbf{Lastly}, we explored the impacts of different prompt types for our proposed framework in ICH segmentation and compared it against state-of-the-art (SOTA) supervised and weakly supervised techniques.

\section{Related Works}
 ICH segmentation methods still primarily rely on fully supervised approaches \cite{9lee2019explainable, 21chang2018hybrid, 20kuo2018cost, 24cho2019improving} and often with in-house datasets. More recently, semi-supervised techniques \cite{semi-supervised-wang2020segmentation} have also been proposed for ICH quantification. However, refined segmentation ground truths are still crucial for their success, and more practical weakly supervised methods are gaining interest. In the limited prior works in this direction, most have relied on categorical labels as weak ground truths. For example, Wu et al. \cite{wu2019weakly} proposed to use refined CAM results and representation learning to achieve ischemic stroke lesion segmentation, achieving a 0.3827 mean Dice score on multi-spectral MRIs. Later, from a binary classification CNN, Nemcek et al. \cite{nemcek2021weakly} detected the location of ICH as bounding boxes in axial brain CT slices using the local extrema of derived attention maps, with a mean Dice of 0.58 for the lesion bounding boxes. Recently, Rasoulian et al. \cite{rasoulian-2023} utilized Head-Wise Gradient-Infused Self-Attention Maps from a Swin Transformer (Swin-HGI-SAM) trained on binary labels (ICH vs. no ICH) to obtain ICH segmentation, which obtained a mean Dice score of 0.438 on CT scans. The recent introduction of SAM \cite{SAM}, which allows interactive prompting in the forms of bounding boxes and/or points for zero-shot segmentation has attracted significant attention. However, its performance on CT-based ICH quantification and as an integrated solution allowing full automation in weakly supervised segmentation is yet to be explored. Furthermore, YOLO models \cite{yolo-review} have been employed for ICH detection \cite{yolo-ich-detection}, but no reports have investigated their potential to facilitate the automation of SAM in ICH segmentation thus far.

\section{Methods and Materials}

\subsection{Dataset and Preprocessing}
For our study, we used the public Brain Hemorrhage Extended (BHX) dataset \cite{BHX}, which includes bounding box annotations for ICH along with their corresponding lesion subtypes, and the manually labeled PhysioNet CT dataset \cite{CNNSegmentation}, which includes manual ICH segmentations. While 4607 CT slices and 5543 bounding boxes from the BHX dataset (containing the ICH subtypes and healthy scans) were employed to train and validate the YOLO model for lesion bounding box detection, the PhysioNet ICH segmentation dataset, which has 2814 CT slices with 318 mask-annotated ICH slices, was reserved as an independent test set to evaluate ICH segmentation with SAM. In addition, for our selected fully supervised baseline methods (more details in Section 3.3), subject-wise five-fold cross-validation was used on the manually segmented PhysioNet dataset to provide segmentation results for all cases and ensure that no slices from the same subject exist across different folds. As CT scans typically have a high dynamic range, for each CT slice, brain, subdural, and bone windows were created based on previous guidelines \cite{RSNA-flanders2020construction} and stacked together to form a composite RGB image, which was normalized to the range of [0,1] in each channel to facilitate training. 

\begin{figure}[t]
  \centering
  \includegraphics[width=\linewidth]{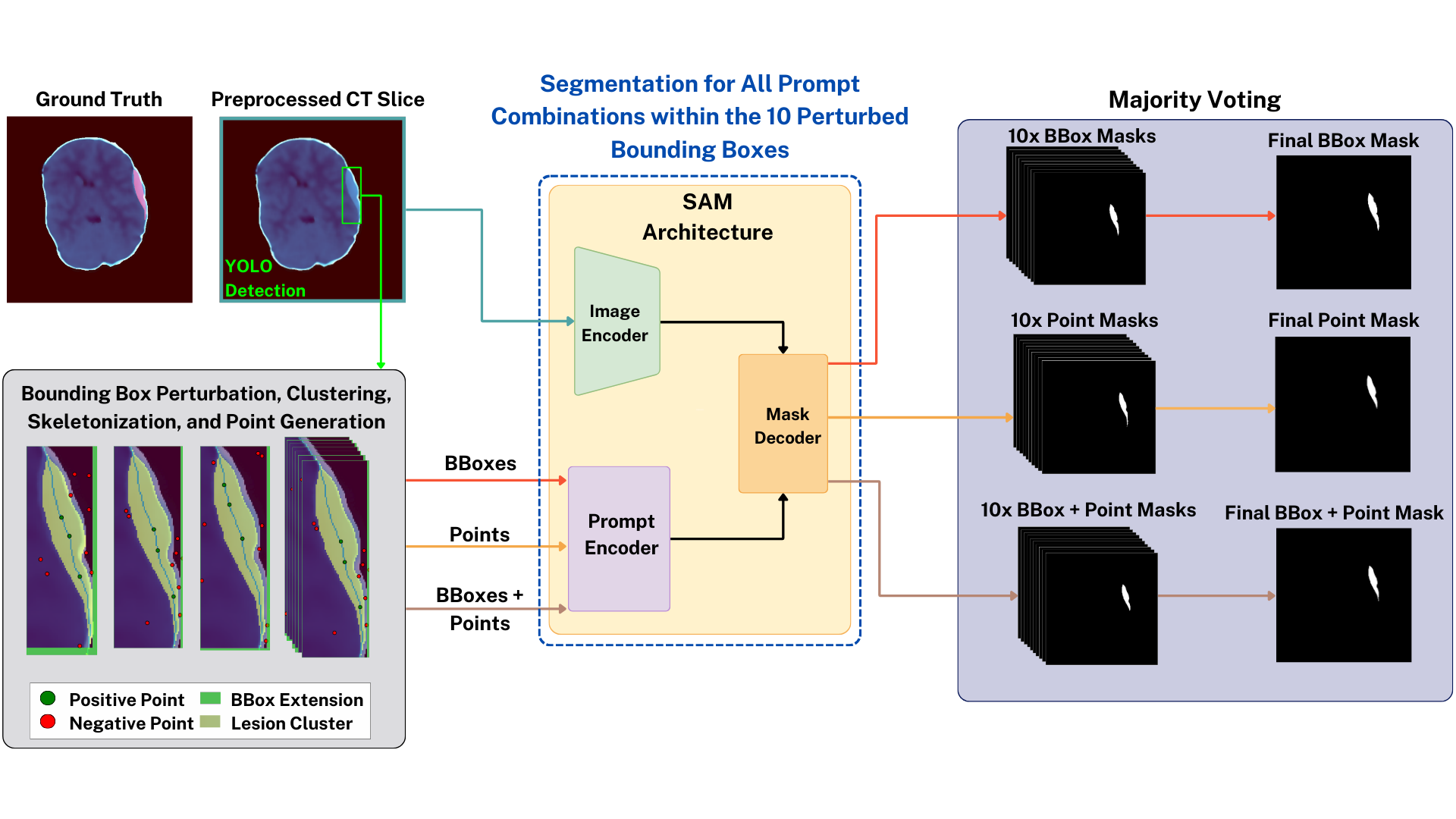} 
  \caption{Workflow of the proposed weakly supervised ICH segmentation method.}
  \label{Workflow}
\end{figure}

\subsection{Uncertainty-Rectified YOLO-SAM Models}
We propose YOLO-SAM, a novel weakly supervised framework for ICH segmentation, where the YOLOv8 model \cite{yolo-review} provides several prompts for SAM to perform ICH segmentation. Here, we built three YOLO-SAM variants, including YOLO-SAM-BBox, YOLO-SAM-Point, and YOLO-SAM-PointBBox, which perform ICH segmentation using bounding box prompts, point prompts, and combinations of bounding boxes and point prompts, respectively. These models each employ an uncertainty rectification strategy that combines 10 SAM outputs based on their 10 respective perturbed prompts. The detailed procedure of our methods is described below and shown in Fig. \ref{Workflow}.

\noindent
\textbf{YOLO Detection:} The preprocessed CT slices are passed to YOLOv8, which outputs the bounding boxes and associated lesion types for detected ICH. Then, the corner coordinates of the predicted bounding boxes are recorded to serve as the basis for automatic point prompt generation.

\noindent
\textbf{Bounding Box Perturbations:} To enhance segmentation robustness and facilitate downstream uncertainty rectification in SAM's outputs, we introduce a method involving bounding box perturbation. Specifically, each YOLO-predicted bounding box is perturbed 10 times by randomly increasing its size by 1-4 pixels on each side. These perturbed boxes are recorded for the next step.

\noindent
\textbf{Clustering and Point Prompt Generation:} Next, to strengthen the prompts' efficacy for SAM, leveraging the predicted ICH bounding box from YOLO, we introduce a novel point prompt generation method for the lesion and background based on a tailored tissue clustering solution and morphological analysis. To delineate lesions in proximity to the skull (e.g. SDH hemorrhage) for SAM, BET \cite{bet} skull-stripping is first applied to the entire CT image. Then, within the ICH bounding box for the skull-stripped RGB composite CT slice, K-means clustering is applied for tissue classification. Here, we use four clusters ($K=4$) regardless of hemorrhage sub-types. If any residual skull is present, in decreasing order of the Hounsfield unit (HU) value, we must account for 1) residual skull tissue 2) ICH 3) healthy brain tissue and 4) dark background; if not, we can expect the brightest cluster to be assigned to the lesion and the remaining 3 clusters to be assigned to the rest. Then, an algorithm is devised to automatically identify the lesion cluster out of the four (YOLO-Clustering). The resulting simple tissue clustering is obtained by first inspecting whether the cluster with the highest average HU value corresponds to the brightest signals in the bone window channel, which represents the residual bone. If not, the cluster is selected as the lesion cluster. Otherwise, the algorithm picks the cluster with the second-highest average HU value. Finally, on the K-means-based lesion clusters, skeletonization is performed to extract the skeleton of the shapes. From these skeletons, positive ICH point prompts are sampled. Then, from each of the three other clusters, negative points are sampled for SAM segmentation. 

\noindent
\textbf{SAM Segmentation with Uncertainty Rectification:} For each of the 10 perturbed bounding boxes, each combination of generated prompts (bounding box, points, and point-box) are passed to SAM's prompt encoder along with the input image to produce a segmentation sample (Fig. \ref{Workflow}). For each YOLO-SAM variant, their final segmentation is obtained via majority voting based on 10 segmentation samples from the associated prompt type. This voting mechanism ensures the robustness of ICH segmentation against network-related prompt instability and SAM's potential sensitivity to these variations, further improving segmentation quality.

\subsection{Baseline Models and Ablation Study}
To validate our proposed method, we compared its performance against the SOTA weakly supervised and fully supervised segmentation techniques for ICH segmentation. With an open-source repository and good performance, we chose the recent Swin-HGI-SAM \cite{rasoulian-2023} as our weakly supervised baseline. In terms of baseline methods with full supervision, we selected the popular UNet \cite{u-net-ronneberger2015u} and Swin-UNETR models \cite{Swin-UNetR-hatamizadeh2021swin}, which have demonstrated strong performance in a wide range of medical image segmentation tasks. For the UNet model, we implemented the architecture from the manually segmented PhysioNet CT data paper \cite{CNNSegmentation}, with four hierarchical layers in the encoding and decoding paths. For the Swin-UNETR model \cite{Swin-UNetR-hatamizadeh2021swin}, we also adopted four hierarchical levels to be consistent with the UNet model.   

While the SAM model \cite{SAM} allows both bounding boxes and/or points as interactive prompts to generate segmentation results, the robustness and accuracy of individual prompt types and their combined usage still require further investigation. Therefore, besides comparison with the baseline models, we also performed an ablation study on the impact of prompt types for the target task (YOLO-SAM-Point, YOLO-SAM-BBox, YOLO-SAM-PointBBox).

\subsection{Model Training \& Evaluation Metrics}
The YOLOv8-m model pretrained on the MS COCO dataset \cite{MSCOCO} in our YOLO-SAM variants was finetuned on the BHX dataset \cite{BHX}, with 3685 CT-slice images and 4479 bounding box labels for the training set, as well as 922 CT-slice images with 1064 labels for the validation set. We used the default YOLOv8 configuration (batch size=16, patience=100) during training. For the first 10,000 iterations, the AdamW optimizer was used with a learning rate of 0.00111 (calculated by a fitting equation using the number of bbox classes, which was 5 for each ICH subtype) and a momentum of 0.9. For the remaining iterations (after epoch 44), the SGD optimizer with an initial learning rate of 0.01 and momentum of 0.9 was used. We trained the weakly supervised Swin-HGI-SAM model \cite{rasoulian-2023} with the RSNA 2019 Brain CT hemorrhage dataset \cite{RSNA-flanders2020construction} (90\%:10\% data split for training vs. validation) following the details from the original publication. As for the supervised baselines (UNet and Swin-UNETR), subject-wise five-fold cross-validation was employed exclusively on the manually segmented PhysioNet dataset, using the AdamW optimizer with an initial learning rate of 0.001 as well as a loss function based on Dice coefficient and cross-entropy. We conducted all model training on a desktop computer with an Intel Core i9 CPU and an NVIDIA GeForce RTX 3090 GPU. \textit{After model training, all evaluations were based on the manually segmented PhysioNet dataset in a slice-wise manner using the default YOLO confidence threshold of 0.25}. As ICH detection is a crucial component of our method, besides segmentation, we also evaluated the binary ICH detection performance (ICH vs. no ICH) for all DL models with accuracy, precision, recall, AUC, F1-score, and specificity. Note that a YOLO prediction was considered a true positive if it correctly identified a slice containing ICH, irrespective of the predicted subtype. For UNet and Swin-UNETR, a true-positive detection was defined as a slice with ICH segmentation that contains more than 10 pixels. In terms of segmentation, we computed the Dice coefficient and Intersection over Union (IoU) for all proposed and baseline models. Paired two-sample t-tests were then used to compare the Dice and IoU scores between the proposed method and the baselines, with $p<0.05$ indicating a statistically significant difference.

\section{Results}

\subsection{Detection Performance}
\begin{table}[b]
\centering
\caption{Detection Performance of Different Methods}
\setlength{\tabcolsep}{5pt} 
\label{table:detection_performance}
\begin{tabular}{lcccc}
\toprule
\textbf{Metric} & \textbf{Swin-HGI-SAM} & \textbf{U-Net} & \textbf{Swin-UNETR} & \textbf{YOLOv8-m} \\
\midrule
\textbf{Accuracy}    & 0.950 & 0.647 & 0.655 & \cellcolor{gray!30}0.933 \\
\textbf{Precision}   & 0.765 & 0.239 & 0.253 & \cellcolor{gray!30}0.665 \\
\textbf{Recall}      & 0.791 & 0.901 & 0.907 & \cellcolor{gray!30}0.626 \\
\textbf{AUC}         & 0.880 & 0.757 & 0.764 & \cellcolor{gray!30}0.796 \\
\textbf{F1-score}    & 0.767 & 0.373 & 0.374 & \cellcolor{gray!30}0.645 \\
\textbf{Specificity} & 0.969 & 0.612 & 0.622 & \cellcolor{gray!30}0.966 \\
\bottomrule
\end{tabular}
\label{table:performance_metrics}
\end{table}

The ICH detection performance for all models is listed in Table \ref{table:detection_performance}. Similar to Swin-HGI-SAM, the YOLOv8-m model demonstrated superior detection performance for most metrics compared to the U-Net and Swin-UNETR models, particularly in precision (0.665 vs. 0.239 and 0.253), AUC (0.796 vs. 0.757 and 0.764), F1-score (0.645 vs. 0.373 and 0.374), and specificity (0.966 vs. 0.612 and 0.622). This highlights the potential of using bounding box localization models such as YOLO to achieve superior performance compared to mask-trained approaches on limited data (UNet, Swin-UNETR) and competitive performance with models trained on substantially more binary labels (Swin-HGI-SAM). However, a weakness of the YOLOv8-m model is its lower slice-wise recall compared to Swin-HGI-SAM (0.626 vs. 0.791), indicating that Swin-HGI-SAM will more reliably detect true positives. For all other detection metrics, YOLOv8-m demonstrated comparable but marginally weaker detection performance, likely due to the smaller number of training samples (4607 bounding box annotated CT slices for YOLO versus 677523 binary-labelled CT slices for Swin-HGI-SAM).

\subsection{Segmentation Performance}

\begin{figure}[h!]
  \centering
  \begin{minipage}{\textwidth}
    \centering
    \includegraphics[width=\linewidth]{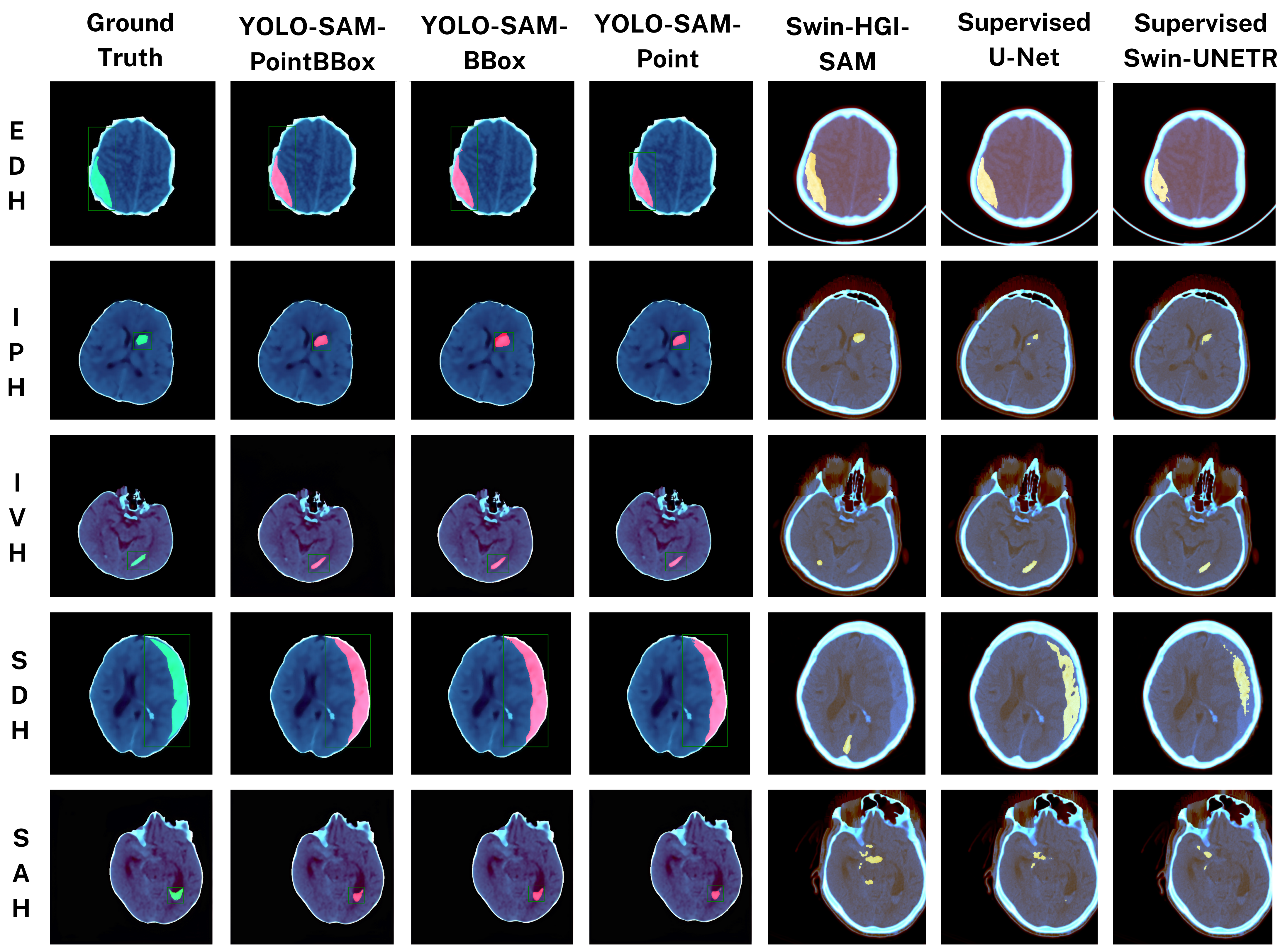} 
    \caption{Qualitative segmentation results on different ICH subtypes}
    \label{qualitative}
  \end{minipage}\\[3ex] 
  \begin{minipage}{\textwidth}
    \centering
    \captionof{table}{Segmentation Performance of Different Models (mean $\pm$ standard error)}
    \setlength{\tabcolsep}{15pt} 
    \begin{tabular}{lcc}
      \toprule
      \textbf{Model} & \textbf{Dice} & \textbf{IoU} \\
      \midrule
      Swin-HGI-SAM & $0.403 \pm 0.014$ & $0.283 \pm 0.011$ \\
      Fully supervised U-Net & $0.388 \pm 0.019$ & $0.297 \pm 0.016$ \\
      Fully supervised Swin-UNETR & $0.428 \pm 0.018$ & $0.330 \pm 0.011$ \\
      YOLO-Clustering & $0.625 \pm 0.020$ & $0.506 \pm 0.019$ \\
      YOLO-SAM-BBox & $0.562 \pm 0.020$ & $0.445 \pm 0.018$ \\
      \textbf{YOLO-SAM-Point} & $\mathbf{0.627 \pm 0.018}$ & $\mathbf{0.506 \pm 0.017}$ \\
      \textbf{YOLO-SAM-PointBBox} & $\mathbf{0.629 \pm 0.018}$ & $\mathbf{0.508 \pm 0.017}$\\
      \bottomrule
    \end{tabular}
    \label{table:segmentation_performance}
  \end{minipage}
\end{figure}

The ICH segmentation results are shown in Table \ref{table:segmentation_performance}, with qualitative outcomes demonstrated in Fig. \ref{qualitative}. Table \ref{table:segmentation_performance} shows that point prompts, hybrid point and bounding box prompts, as well as simple tissue clustering within the YOLO bounding box (YOLO-Clustering) yielded significantly higher segmentation performance than Swin-HGI-SAM, UNet, Swin-UNETR and using bounding box prompts alone ($p<0.005$). It is also shown that hybrid prompts have improved performance over point prompts and YOLO-Clustering on average, though not statistically significant ($p>0.05$). While YOLO-Clustering had good segmentation quality, it also had higher standard error than SAM with hybrid and point prompts, highlighting the point prompt's better precision and reliability. Finally, while YOLO-SAM-BBox does not show significantly higher Dice and IoU scores than UNet ($p=0.0853$) or Swin-UNETR ($p=0.768$), it significantly outperforms Swin-HGI-SAM ($p<0.005$).

\vspace{-0.4cm}

\section{Discussion}
Our YOLO-SAM framework that integrates YOLOv8-m, a novel point-prompt generator, and SAM with uncertainty rectification has demonstrated great performance in weakly supervised ICH segmentation, particularly with the hybrid prompts. The superior performance over existing weakly supervised and fully supervised methods can be explained by the incorporation of the power of the foundation models and spatial information represented by the bounding box ground truths. It is important to acknowledge that the poor performance of fully supervised DL models, such as UNet and Swin-UNETR can also be partially due to the low number of ground-truth mask labels. Despite this success, the slice-wise recall metric for our YOLO model lagged behind the Swin-HGI-SAM, suggesting a potential compromise in the model's ability to detect all ICH slices. However, after investigating this further on a patient-wise basis, the recall metric was calculated at 0.9714, with 34 out of 35 patients with hemorrhage having had at least one slice detected. In a clinical setting, the proportion of true positive ICH cases would therefore be much higher than the reported slice-wise recall metric. Our ablation study showed that hybrid prompts offered better performance than points or bounding boxes. This observation echoes previous reports \cite{medsam-ma2024segment} and could be explained by the lack of robustness when capturing thin, elongated, and curved structures (e.g., IPH subtype) with bounding boxes by SAM. Finally, while MedSAM \cite{medsam-ma2024segment} has gained great popularity in the community, its adoption in our YOLO-MedSAM-BBox model resulted in inferior segmentation outcomes (Dice= 0.412±0.018, IoU=0.298±015). This is consistent with other reports of SAM outperforming MedSAM on certain medical image segmentation tasks \cite{SAM-vs-MedSAM-1-liu2024wsi} and may be due to a lack of public datasets for training MedSAM on ICH tasks, as Dice loss was used in training the model \cite{medsam-ma2024segment}. 

\section{Conclusion}
In conclusion, we have proposed a novel weakly supervised ICH segmentation technique that uses YOLO and an uncertainty-rectified SAM. In addition to bounding boxes provided via YOLO, our morphology-based point prompt generation was proven to offer enhanced segmentation performance. Thorough assessments have revealed its superior performance over SOTA weakly supervised and fully supervised baselines while maintaining strong ICH detection capabilities.   

\begin{credits}
\subsubsection{\ackname} We acknowledge the support of the Natural Sciences and Engineering Research Council of Canada (596537, RGPIN/05100-2022) and the Fonds de recherche du Québec \textbf{–} Nature et technologies (2022-PR296459, B1X-348625).

\subsubsection{\discintname}
The authors declare no competing interests.
\end{credits}
%
%
%
%
\bibliography{Paper-0004}
\bibliographystyle{splncs04}
\end{document}